\pdfoutput=1

\documentclass[11pt]{article}

\usepackage{acl}

\usepackage{times}
\usepackage{latexsym}

\usepackage[T1]{fontenc}

\usepackage[utf8]{inputenc}

\usepackage{microtype}

\usepackage{inconsolata}

\usepackage{graphicx}

\usepackage{multirow}
\usepackage{amsmath,amssymb,amsfonts}
\usepackage{amsthm}
\usepackage{mathrsfs}
\usepackage{xcolor}
\usepackage{textcomp}
\usepackage{manyfoot}
\usepackage{booktabs}
\usepackage{algorithm}
\usepackage{algorithmicx}
\usepackage{algpseudocode}
\usepackage{listings}
\usepackage{hyperref}       
\usepackage{url}            
\usepackage{nicefrac}       
\usepackage{microtype}      
\usepackage{cleveref}
\usepackage{caption}
\usepackage[nolist]{acronym}
\usepackage[acronym]{glossaries}
\usepackage{siunitx}
\usepackage{longtable}
\usepackage{enumitem}       
\usepackage{wrapfig}        
\usepackage[font={small}]{caption}
\usepackage{comment}
\usepackage{framed}
\usepackage{xltabular}
\usepackage{chngcntr}

\title{From Citations to Criticality:\\Predicting Legal Decision Influence in the Multilingual Swiss Jurisprudence}

\author{
Ronja Stern $^{1}$\thanks{\hspace{2mm} Equal contribution.}
\And
Ken Kawamura $^{5 *}$
\AND
Matthias Stürmer $^{1,2}$
\And
Ilias Chalkidis $^{4}$
\And
Joel Niklaus $^{1,2,3 *}$
\AND
\\
$^1$University of Bern\quad
$^2$Bern University of Applied Sciences\\
$^3$Stanford University
$^4$University of Copenhagen\quad
$^5$Independent Scholar\quad
}

\begin{document}
\maketitle

\begin{acronym}[UMLX]
    \acro{FSCS}{Federal Supreme Court of Switzerland}
    \acro{SCI}{Supreme Court of India}
    \acro{ECHR}{European Convention of Human Rights}
    \acro{ECtHR}{European Court of Human Rights}
    \acro{SCOTUS}{Supreme Court of the United States}
    \acro{SPC}{Supreme People's Court of China}
    \acro{SJP}{Swiss-Judgment-Prediction}
    \acro{ASO}{Almost Stochastic Order}
    \acro{ILDC}{Indian Legal Documents Corpus}
    
    \acro{US}{United States}
    \acro{EU}{European Union}

    \acro{NLP}{Natural Language Processing}
    \acro{ML}{Machine Learning}
    \acro{LJP}{Legal Judgment Prediction}
    \acro{SJP}{Swiss-Judgment-Prediction}
    \acro{PJP}{Plea Judgment Prediction}
    
    \acro{BERT}{Bidirectional Encoder Representations from Transformers}
    \acro{LSTM}{ Long Short-Term Memory }
    \acro{GRU}{Gated Recurrent Unit}
    \acro{BiLSTM}{Bidirectional Long Short-Term Memory}
    \acro{CNN}{Convolutional Neural Networks}

    \acro{PLM}{pre-trained Language Model}
    \acro{LM}{Language Model}

    \acro{RTD}{Replaced Token Detection}

    \acro{CLT}{Cross-Lingual Transfer}
    \acro{HRL}{high resource language}
    \acro{LRL}{low resource language}

    \acro{POS}{Part-of-Speech}
    
    \acro{SLTC}{Single Label Text Classification}
    \acro{MLTC}{Multi Label Text Classification}
    \acro{TC}{Text Classification}
    \acro{NLU}{Natural Language Understanding}
    \acro{IR}{Information Retrieval}
    \acro{NER}{Named Entity Recognition}
    \acro{NLU}{Natural Language Understanding}
    \acro{QA}{Question Answering}
    \acro{NLI}{Natural Language Inference}

    \acro{GNB}{Gaussian Naive Bayes}
    \acro{DT}{Decision Tree}
    \acro{SVM}{Support-Vector Machine}
    \acro{RF}{ Random Forest}
    \acro{XGBoost}{eXtreme Gradient Boosting}
    \acro{MLIR}{Multilingual Information Retrieval}
    \acro{IR}{Information Retrieval}
    \acro{NDCG}{Normalized Discounted Cumulative Gain}
    \acro{LD}{Leading Decision}
    \acro{FSCD}{Federal Supreme Court Decisons}
    \acro{SFCS}{Swiss Federal Supreme Court}
    \acro{LLM}{Large Language Model}
    \acro{LM}{Language Model}

    \acro{CVG}{Court View Generation}
    \acro{JP}{Judgment Prediction}
    \acro{LAP}{Law Area Prediction}
    \acro{SLAP}{Sub Law Area Prediction}
    \acro{CP}{Criticality Prediction}
    \acro{LDS}{Leading Decision Summarization}
    \acro{CE}{Citation Extraction}
    
    \acro{SBERT}{Sentence-Bert}
    \acro{Doc2Doc}{Document-to-Document}
    \acro{mUSE}{Multilingual Universal Sentence Encoder}

\end{acronym}

\begin{abstract}
Many court systems are overwhelmed all over the world, leading to huge backlogs of pending cases. Effective triage systems, like those in emergency rooms, could ensure proper prioritization of open cases, optimizing time and resource allocation in the court system.
In this work, we introduce the Criticality Prediction dataset, a novel resource for evaluating case prioritization.  
Our dataset features a two-tier labeling system: (1) the binary LD-Label, identifying cases published as Leading Decisions (LD), and (2) the more granular Citation-Label, ranking cases by their citation frequency and recency, allowing for a more nuanced evaluation.
Unlike existing approaches that rely on resource-intensive manual annotations, we algorithmically derive labels leading to a much larger dataset than otherwise possible. We evaluate several multilingual models, including both smaller fine-tuned models and large language models in a zero-shot setting. Our results show that the fine-tuned models consistently outperform their larger counterparts, thanks to our large training set.
Our results highlight that for highly domain-specific tasks like ours, large training sets are still valuable.

\end{abstract}

\section{Introduction}
Predicting the impact of legal cases is a critical task in the legal domain, as it aids professionals in the judicial system prioritize and manage large volumes of case law more effectively, ensuring that critical cases receive the necessary attention.
Despite its significance, the task of predicting the case criticality remains relatively under-explored. Existing approaches to evaluating the importance of legal cases are primarily manual, very resource-intensive and subject to the judgments of individual annotators. This paper introduces a novel dataset
, licensed under CC~BY~4.0,\footnote{\url{https://huggingface.co/datasets/rcds/swiss_criticality_prediction}}
and a more challenging evaluation framework—\textbf{Criticality Prediction}—designed to predict the potential influence of Swiss Federal Court cases on future jurisprudence.

While prior work such as the Importance Prediction task for European Court of Human Rights (ECtHR) \cite{chalkidis_neural_2019} cases focused on predicting importance on a defined scale using human-assigned labels, our approach employs algorithmically derived labels to evaluate case criticality. Our dataset introduces a two-tier labeling system: the \textbf{LD-Label}, a binary indicator of whether a case is published as a Leading Decision (LD), and the \textbf{Citation-Label}, a more nuanced categorization based on the frequency and recency-weighted citation counts of these decisions across subsequent cases. This distinct formulation of "criticality" not only distinguishes critical cases from non-critical ones but also ranks them by their relative importance over time. The complexity of this dataset challenges even recent LLMs such as ChatGPT.

Our contributions are threefold: (1) We propose a novel Criticality Prediction task that provides a more comprehensive and challenging evaluation of case law importance. (2) We release the datasets to the community, providing valuable resources for further research in legal NLP. (3) We evaluate several multilingual models of various sizes, including fine-tuned variants to establish baselines.


\begin{figure}
    \centering
    \includegraphics[width=\linewidth]{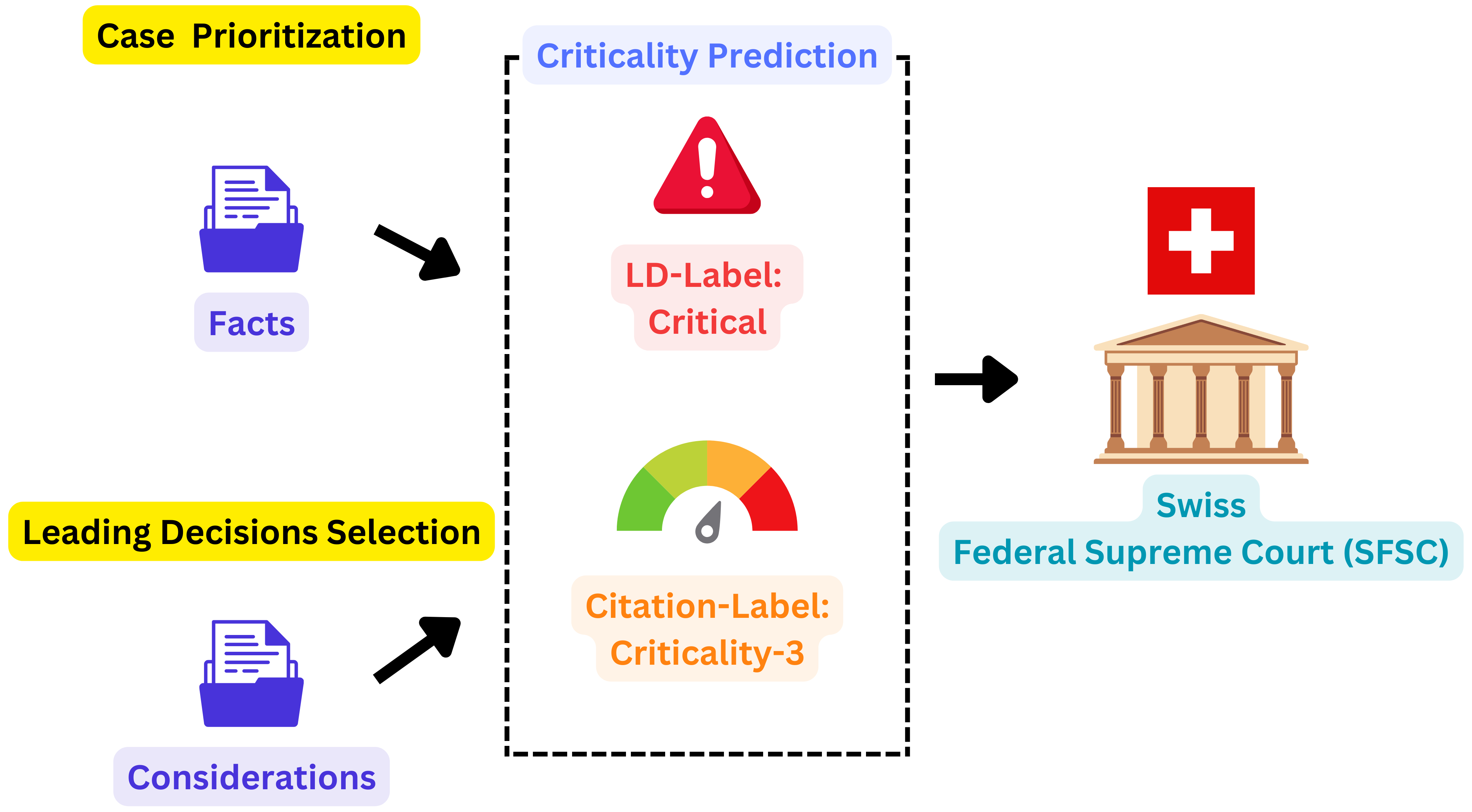}
    \caption{Overview of the Criticality Prediction Task.}
    \label{fig:task-overview}
\end{figure}
\section{Related Work}

    One of the most common text classification tasks in the legal domain is Legal Judgment Prediction (LJP), which involves predicting outcome of a case based on its facts. Researchers have leveraged datasets with unique characteristics to analyze and predict case outcomes across various languages, jurisdictions, and input types~\citep{ijcai2022p765, aletras_predicting_2016, sulea_predicting_2017, medvedeva_judicial_2018, chalkidis_neural_2019, niklaus_swiss-judgment-prediction_2021, niklaus_empirical_2022,  semo_classactionprediction_2022}. While LJP is focused on the outcome of individual cases, Importance Prediction shifts the focus toward assessing the broader significance of a case.

    \citet{chalkidis_neural_2019} introduced the Importance Prediction task using cases from the European Court of Human Rights (ECtHR). In this task, ECtHR provided scores that denote each case’s `importance' to the common law. These scores, ranging from 1 (key case) to 4 (unimportant), were designed to help legal practitioners identify cases that play a crucial role in shaping jurisprudence. The labels reflect the long-term impact of a case on future rulings and the evolution of legal precedent. While the task is invaluable for identifying landmark cases, it relies on legal experts to assign the labels, making the process resource-intensive and potentially subject to subjective interpretations. Additionally, the \mbox{ECtHR} dataset contains only English samples, whereas our Swiss dataset is multilingual. To our knowledge, no other study has specifically addressed the task of Criticality Prediction.


    
\section{Task and Dataset}
\label{task_and_dataset}

\begin{table*}[ht]
    \centering
        \caption{Task Configurations. Label names are \emph{Critical} (C), \emph{Non-critical} (NC), \emph{Critical-1} (C1) to \emph{Critical-4} (C4).}
    \footnotesize
    \resizebox{\textwidth}{!}{%
    \begin{tabular}{lrrrrrrrrrrrrrrr}
         \toprule
         \textbf{Task Name} & \multicolumn{5}{c}{\textit{Train}} & \multicolumn{5}{c}{\textit{Validation}}& \multicolumn{5}{c}{\textit{Test}} \\
         \cmidrule(lr){2-6}\cmidrule(lr){7-11}\cmidrule(lr){12-16}
          & \multicolumn{1}{c}{\textbf{Size}} & \multicolumn{4}{c}{\textbf{Label Distribution}} & \multicolumn{1}{l}{\textbf{Size}} & \multicolumn{4}{c}{\textbf{Label Distribution}}& \multicolumn{1}{c}{\textbf{Size}} &  \multicolumn{4}{c}{\textbf{Label Distribution}}\\
         \midrule
            && \textbf{C} & \textbf{NC} &&&& \textbf{C} & \textbf{NC} &&&& \textbf{C} & \textbf{NC} \\
            LD-Facts  & \textbf{74799} & 2542 & 72257 &-&-& \textbf{12019} & 580 & 11439 &-&-& \textbf{26239} & 950 & 25289&-&-\\
            LD-Considerations & \textbf{87555} & 2544 & 85011 &-&-& \textbf{13386} & 580 & 12806 &-&-& \textbf{29486} & 948 & 28538&-&-\\
        \midrule
           && \textbf{C-1} & \textbf{C-2} & \textbf{C-3} & \textbf{C-4} &&\textbf{C-1} & \textbf{C-2} & \textbf{C-3} & \textbf{C-4} && \textbf{C-1} & \textbf{C-2} & \textbf{C-3} & \textbf{C-4} \\
            Citation--Facts  & \textbf{2506} &782 & 626 & 585 & 513 & \textbf{563} &186 & 152 & 131 & 94 & \textbf{725} & 137 & 177 & 224 & 187\\
            Citation-Considerations  & \textbf{2509}&779 & 624 & 586 & 520 & \textbf{563} &186 & 154 & 131 & 92 & \textbf{723} &137 & 177 & 224 & 185\\
        \bottomrule
    \end{tabular}
    }
    \label{tab:lextreme-task-dist}
\end{table*}

\subsection{Criticality}
\label{sec:cp}
Understanding the legal framework of the Swiss Federal Supreme Court (SFSC) is key to defining criticality in Swiss case law. The SFSC shapes the legal landscape through its rulings, with a subset known as \acp{LD} published separately due to their influence on the interpretation of law. We quantify case criticality using two labels: the \textbf{LD-Label} and the \textbf{Citation-Label}.

The \textbf{LD-Label} is binary, categorizing cases as \emph{critical} or \emph{non-critical}. SFSC cases are labeled \emph{critical} if also published as \ac{LD}, reflecting their recognized importance within the Swiss legal system. We used regular expressions to extract SFSC cases published as LD. SFSC cases not published as LD are labeled \emph{non-critical}.

We developed the \textbf{Citation-Label} for a more granular measure. It counts how often each \ac{LD} case is cited in SFSC cases, with less weight on older cases to prioritize recency. The score is calculated as: \begin{math}score=count\times\frac{year-2002+1}{2023-2002+1}\end{math}. The \textit{count} is the citation frequency, and the weighting reduces older cases' influence. This score ranks \ac{LD} cases, and we categorized them into four criticality levels—\emph{critical-1} (least critical) to \emph{critical-4}—based on the 25th, 50th, and 75th percentiles. More details on the constants are in Appendix~\ref{sec:appendix-weighting}.

\begin{figure}
    \captionsetup{skip=4pt}
    \centering
    \includegraphics[width=\columnwidth]{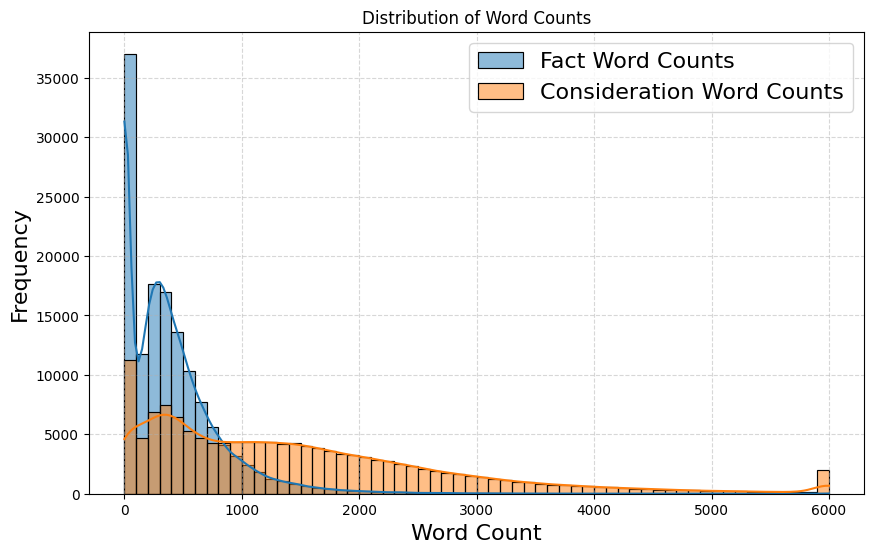}
    \caption{Facts (\textcolor{cyan}{Blue}) and consideration (\textcolor{orange}{Orange}) length distribution measured in Words with Spacy~\cite{honnibal_spacy_2020}.  Those with more than 6000 words are binned together.}
    \label{fig:cp_facts_length_distribution}
    \label{fig:cp_considerations_length_distribution}
    \label{fig:enter-label}
\end{figure}

Unlike a prior approach~\cite{chalkidis_neural_2019}, which also used four categories, we explicitly incorporate temporal weighting to account for both the influence of a case and how its criticality shifts over time. Furthermore, our framework allows for the dynamic recalculation of scores and re-labeling of criticality as case law evolves. This adaptability ensures that our dataset reflects the ongoing changes in the legal system, whereas prior studies would require manual re-annotation as the law develops.



\subsection{Criticality Prediction Task and Dataset}

The \ac{CP} task leverages two primary inputs from the \ac{SFCS} cases: \textbf{facts} and \textbf{considerations}. Facts describe a factual account of the events of each case and form the basis for the considerations of the court. Considerations reflect the formal legal reasoning, citing laws and other influential rulings, and forming the basis for the final ruling.

We see two distinct applications for these inputs in the Criticality Prediction task, illustrated in Figure~\ref{fig:task-overview}. In the \emph{Case Prioritization} task, only the facts are used as input. This produces a score indicating how critical or important a case is. The goal is to help prioritize cases, which could assist in determining which cases should be heard sooner or assigned to more experienced judges. Many court systems are overwhelmed all over the world leading to huge backlogs of pending cases. Effective triage systems, like those in emergency rooms, could ensure proper prioritization of open cases.

In the \emph{Leading Decisions Selection} task, considerations are used for a post-hoc analysis, comparing the ruling to prior case law to assess its potential impact on future jurisprudence. This analysis is exactly what the Supreme Court does at the end of the year to arrive at the selection of leading decisions.

Our approach, by offering facts and considerations as inputs, reflects different stages of legal processing. Both tasks can utilize the LD-Label or Citation-Label, providing varying levels of granularity. This dual approach enhances the dataset’s practical utility, supporting both early-stage prioritization and post-judgment evaluation, thereby addressing multiple aspects of legal workflows.

Our dataset spans from 2002 to 2023, and is partitioned into train (2002-2015), dev (2016-2017), and test (2018-2022) sets, as outlined in Table~\ref{tab:lextreme-task-dist}. We allocated a relatively large test set to accommodate longitudinal studies, including the COVID-19 pandemic years. This setup ensures that the test set represents the most recent data for realistic evaluation \cite{sogaard_we_2021}. The dataset consists of 138,531 total cases, with 85,167 in German, 45,451 in French, and 7,913 in Italian. See \Cref{fig:cp_facts_length_distribution} for length distributions of facts and considerations and Appendix~\Cref{tab:metadata} for general dataset metadata.

\section{Methods}

\subsection{Models}
We evaluated the following models: XLM-R (Base and Large)~\citep{conneau_cross-lingual_2019}, MiniLM~\citep{wang2020minilm}, DistilmBERT~\citep{sanh2019distilbert}, mDeBERTa-v3~\citep{he2021debertav3}, X-MOD (Base)~\citep{pfeiffer_lifting_2022}, SwissBERT~\citep{vamvas_swissbert_2023}, mT5 (Small and Base)~\citep{xue_mt5_2021}, BLOOM (560M)~\citep{scao_bloom_2022}, Legal-Swiss-RoBERTa and Legal Swiss Longformer (Base)~\citep{Rasiah2023OneLM}, GPT-3.5~\citep{brown_language_2020}, and LLaMA-2~\citep{touvron_llama_2023}.

We fine-tuned all models per task, using early stopping on the dev set. Due to resource constraints, further fine-tuning of GPT-3.5 and LLaMA-2 is reserved for future work, with their current performance serving as baseline results. SwissBERT and Legal-Swiss models were chosen for their Swiss-specific pretraining, while the other models were chosen for their multilingual capabilities, essential for our multilingual dataset. 

We evaluated GPT-3.5 and LLaMA-2 in a zero-shot setting following \citet{chalkidis2023chatgptpassbarexam}. Samples were randomly selected from the validation set to prevent test set leakage for future evaluations especially for a closed model (GPT-3.5). To manage costs, we limited the validation set to 1000 samples. Our experiments focused on zero-shot classification due to the long input lengths. The prediction labels were determined using regular expressions to match model outputs to class labels. We show the prompts used in Appendix Figure~\ref{fig:CP-LD-Facts} and Figure~\ref{fig:prompt-citation}. 

\subsection{Metrics}

We adopt the LEXTREME benchmark setup \citep{niklaus-etal-2023-lextreme}, and use aggregation of macro-averaged F1 scores with the harmonic mean to emphasize lower scores, promoting fairness across languages and input types. Scores are averaged over random seeds, languages (de, fr, it), and input types (facts or considerations), encouraging consistent performance across configurations.

\section{Results}
\begin{table*}[ht]
    \centering
    \caption{Configuration aggregate scores. The macro-F1 scores from the language-specific subsets of the test set are provided. 
    }
    \footnotesize
    \resizebox{\textwidth}{!}{%
    \begin{tabular}{lccccccccc}
\toprule
\bf{Model} &          \bf{LD-F} &          \bf{LD-C} &          \bf{C-F} &          \bf{C-C} &  \bf{Agg.} \\
Languages & \hspace{1.2mm}de\hspace{1.3mm} / \hspace{1.2mm}fr\hspace{1.3mm} / \hspace{1.2mm}it\hspace{1.3mm}& \hspace{1.2mm}de\hspace{1.3mm} / \hspace{1.2mm}fr\hspace{1.3mm} / \hspace{1.2mm}it\hspace{1.3mm}& \hspace{1.2mm}de\hspace{1.3mm} / \hspace{1.2mm}fr\hspace{1.3mm} / \hspace{1.2mm}it\hspace{1.3mm}& \hspace{1.2mm}de\hspace{1.3mm} / \hspace{1.2mm}fr\hspace{1.3mm} / \hspace{1.2mm}it\hspace{1.3mm}& \hspace{1.2mm}de\hspace{1.3mm} / \hspace{1.2mm}fr\hspace{1.3mm} / \hspace{1.2mm}it\hspace{1.3mm}\\
\midrule
MiniLM                &  57.5 / 53.9 / 52.9 &  68.1 / 65.4 / 64.2 &   12.1 / 13.1 / 6.8 &  24.6 / 21.9 / 17.3 &  25.7 / 25.7 / 16.7 \\
DistilmBERT            &  56.3 / 55.6 / 56.8 &  67.8 / 63.9 / 64.7 &  20.2 / 18.2 / 20.7 &  22.6 / 21.6 / 22.2 &  31.7 / 29.7 / 31.6 \\
mDeBERTa-v3           &  57.6 / 55.1 / 52.7 &  73.9 / 68.1 / 67.7 &  25.4 / 22.8 / 16.8 &  22.1 / 21.6 / 12.6 &  34.6 / 32.5 / 23.2 \\
XLM-R\textsubscript{Base}            &  59.4 / 56.3 / 56.0 &  70.2 / 65.4 / 62.5 &  20.0 / 20.6 / 23.5 &  26.5 / 22.1 / 23.1 &  33.7 / 31.5 / 33.4 \\
XLM-R\textsubscript{Large}           &  58.4 / 56.8 / 54.1 &  70.5 / 67.3 / 66.0 &  22.5 / 19.7 / 36.2 &  26.7 / 28.2 / 33.0 &  46.9 / 33.7 / 43.7 \\
X-MOD\textsubscript{Base}            &  59.0 / 56.2 / 54.8 &  71.1 / 68.7 / 64.1 &  19.8 / 17.2 / 24.4 &  23.2 / 24.2 / 16.4 &  32.1 / 30.3 / 29.5 \\
SwissBERT\textsubscript{(xlm-vocab)} &  57.6 / 55.9 / 57.3 &  72.4 / 69.3 / 61.2 &  23.8 / 20.3 / 39.4 &  28.5 / 24.0 / 18.7 & 36.9 / 32.3 / 35.5 \\
\midrule
mT5\textsubscript{Small} & 54.8 / 51.7 / 50.3 & 69.2 / 61.9 / 56.4 & 14.2 / 16.2 / 10.5 & 15.9 / 18.1 / 20.2  & 24.1 / 26.2 / 21.9 \\
mT5\textsubscript{Base} & 54.1 / 52.1 / 50.3 & 66.4 / 61.9 / 56.8 & 10.6 / 16.3 / 16.9 & 18.7 / 18.7 / 22.1  & 22.1 / 26.6 / 28.2 \\
\midrule
BLOOM-560m & 55.1 / 53.2 / 50.9 & 64.6 / 65.3 / 56.2 &  12.6 / 16.1 / 7.1 &   9.5 / 13.6 / 5.1 & 18.3 / 23.6 / 10.7 \\
\midrule
Legal-Swiss-R\textsubscript{Base}       &  59.3 / 58.4 / 55.5 &  73.8 / 69.4 / 68.6 &  24.3 / 20.5 / 10.5 &  26.2 / 25.3 / 14.0 & 36.5 / 33.4, 20.1   \\
Legal-Swiss-R\textsubscript{Large}      &  58.3 / 55.7 / 53.6 &  71.9 / 68.5 / 66.7 &  23.0 / 21.3 / 38.7 &   28.5 / 26.0 / 9.0 &  36.5 / 33.9 / 23.4 \\
Legal-Swiss-LF\textsubscript{Base}      &  60.7 / 58.3 / 55.5 &  74.8 / 70.0 / 67.8 &  25.3 / 21.5 / 18.4 &   29.2 / 26.7 / 9.9 & 38.6 / 34.7 / 21.3 \\
\bottomrule
\end{tabular}
}
\label{tab:config_aggregate_scores_language_specific}
\end{table*}

\begin{table}[ht]
  \centering
    \caption{Results using Macro F1, with highest values in \textbf{bold}.
        'LD' and 'C' denote LD and Citation labels, while 'F' and 'C' refer to inputs from Facts or Considerations. Models marked with (*) are zero-shot LLMs evaluated on the validation set. 
        }
  \footnotesize
  \resizebox{\columnwidth}{!}{
  \begin{tabular}{lrrrrrr}
    \toprule
        \bf{Model} & \bf{LD-F} & \bf{LD-C} & \bf{C-F} & \bf{C-C} & \bf{Agg.} \\
    \midrule
    Random Baseline & 36.2 & 36.0 & 24.1 & 25.7 & 29.5\\ 
    Majority Baseline& 49.1 & 49.2 & 11.8 & 11.8 & 19.0\\
    \midrule
        MiniLM                &       54.7 &       65.8 &        9.8 &       20.8 & 21.8 \\
        DistilmBERT            &       56.2 &       65.4 &       19.6 &       22.1 &       30.9 \\
        mDeBERTa-v3           &       55.1 &       69.8 &       21.0 &       17.5  &       29.1 \\
        XLM-R\textsubscript{Base}            &       57.2 &       65.9 &       21.3 &       23.7 & 32.8 \\
        XLM-R\textsubscript{Large}           &       56.4 &       67.9 &       24.4 &  \bf{29.1} & \textbf{37.1} \\
        X-MOD\textsubscript{Base}            &       56.6 &       67.8 &       20.0 &       20.6 & 30.5 \\
        SwissBERT\textsubscript{(xlm-vocab)} &       56.9 &       67.3 &       25.7 &       23.0 & 34.8 \\
    \midrule
        mT5\textsubscript{Small}      &       52.2 &       62.1 &       13.2 &       17.9 & 24.0 \\
        mT5\textsubscript{Base}       &       52.1 &       61.5 &       14.0 &       19.7 & 25.4 \\
    \midrule
        BLOOM\textsubscript{560M}            &       53.0 &       61.7 &       10.7 &        8.0 &  15.8 \\
    \midrule
        Legal-Swiss-RoBERTa\textsubscript{Base}       &       57.7 &       70.5 &       16.2 &       20.1 &      28.0 \\
        Legal-Swiss-RoBERTa\textsubscript{Large}      &       55.9 &       68.9 &  \bf{25.8} &       16.3 & 30.2 \\
        Legal-Swiss-LF\textsubscript{Base}      &  \bf{58.1} &  \bf{70.8} &       21.4 &       17.4 & 29.5\\
    \midrule
        GPT-3.5*      &       46.6 &       44.8 &  25.7 &       16.7 &        28.1 \\
        LLaMA-2*       &       45.2 &       26.6 &  7.0 &       8.5 &       12.5 \\
    \bottomrule
    \end{tabular}
    }
    \label{tab:text-clasification-results}
\end{table}

We present results in \Cref{tab:text-clasification-results}, with standard deviations in Appendix \Cref{tab:config_aggregate_scores_with_std} and scores on the validation dataset in Appendix \Cref{tab:config_aggregate_scores_of_validation_set}. The best performance was achieved by XLM-R\textsubscript{Large}, with an aggregate (Agg.) score of 37.1. SwissBERT also demonstrated competitive results, with an Agg. score of 34.8. Interestingly, larger models did not always outperform their smaller counterparts. For example, mT5\textsubscript{Base} and mT5\textsubscript{Small} both underperformed DistilmBERT in all configurations.
The Legal-Swiss models performed well in LD labels, particularly the Legal-Swiss-LF\textsubscript{Base}, which achieved the highest scores in LD-F and LD-C. However, their weak performance on Citation labels highlights the dataset's complexity, even for domain-specific models.
GPT-3.5 and LLaMA-2 underperform fine-tuned models, underlining the need for specialized models for these tasks.
The difference is largest in the LD labels where the small fine-tuned models always outperformed LLMs despite their huge parameter number difference (e.g., MiniLM with 118M parameters versus GPT-3.5 with 175B).


Table~\ref{tab:config_aggregate_scores_language_specific} shows more detailed results on the language specific scores.
SwissBERT pretrained with a focus on German achieved  the highest aggregate score in German, but interestingly with scores in Italian being the highest by far in C-F. 
Models pretrained on CC100~\citep{conneau-etal-2020-unsupervised} (MiniLM, mDeBERTa, XLM-R and X-MOD) exhibited mixed results in French and Italian, with all models performing best in German. MiniLM, mDeBERTa, and X-MOD showed underperformance in Italian but stronger results in French. In contrast, XLM-R, particularly the large variant, demonstrated robust performance in Italian.
mT5 models performed well in French, and the base variant additionally also performed well on Italian. 
BLOOM was much better in French than in other languages, not surprising given it did not have German and Italian in the pretraining data. 

Overall, the connection between the proportion of a language in the pretraining corpus and the model's downstream performance in that language appears weak. While some models showed stronger results in languages with higher representation in their pretraining data, this trend was inconsistent across all models and languages.
\section{Conclusions and Future Work}
This work introduced a novel Criticality Prediction task to assess the potential influence of SFSC cases on future jurisprudence. Our approach utilizes leading decision citation for a more comprehensive and challenging multilingual evaluation of case law importance compared to existing methods. We also released its multilingual dataset for a community.

We conducted a comprehensive evaluation of our Criticality Prediction task, comparing a range of models, from smaller multilingual models to LLMs like GPT-3.5. Our findings show that small fine-tuned models consistently outperform zero-shot LLMs, highlighting the importance of task-specific adaptation in legal NLP applications.

Future studies could apply the Criticality Prediction task in other legal contexts by incorporating sources from different jurisdictions and languages. This approach would broaden the research’s impact and enhance the model’s adaptability across different legal systems.

\section*{Limitations}


It is very difficult to estimate the importance of a case. By relying on proxies such as whether the case was converted to a leading decision (LD-label) and how often this leading decision was cited (Citation-label), we were able to create labels algorithmically. While we discussed this with lawyers at length and implemented the solution we agreed on finally, this task remains somewhat artificial. Additionally, it is worth noting that LLMs can be sensitive to prompt formats and the order of answer options, leading to inconsistent outputs as highlighted in recent research~\cite{webson-pavlick-2022-prompt,elazar-etal-2021-measuring}. Incorporating varied prompt designs into future studies could strengthen our approach further.



\section*{Ethics Statement}
While automating case prioritization and identifying leading decisions can greatly benefit legal professionals, there are potential risks associated with deploying such classifiers. One concern is the risk of perpetuating biases present in historical legal trends. For instance, case prioritization decisions should not be influenced by factors such as gender, race, or other protected characteristics. We acknowledge these concerns and will pursue measures to mitigate such biases in future work.

Additionally, there are challenges with reproducibility when using closed models like ChatGPT. Since the internal workings of these models are not fully transparent, results may be difficult to replicate. To promote open science, we have provided comprehensive evaluations of open source multilingual models, aiming to make our findings more accessible and reproducible.

\section*{Acknowledgments}
We thank Luca Rolshoven and Mara Häusler for their thoughtful feedback on the manuscript, and the anonymous reviewers for their insightful and constructive comments.

\bibliography{references,custom}

\appendix
\clearpage
\section{Weighting Formula for Citation-Label}
\label{sec:appendix-weighting}

The weighting formula used for the \textbf{Citation-Label} is designed to balance the impact of older cases with more recent rulings, ensuring that the ranking reflects both citation frequency and recency. The formula is as follows:

\[ score=count\times\frac{year-2002+1}{2023-2002+1} \]

Where:

\begin{itemize}
    \item \textit{count} refers to the number of times a particular case is cited in Swiss Federal Supreme Court (SFSC) decisions.
    \item The year 2002 is the starting year of our dataset.
    \item The year 2023 is the end point of the our dataset.
\end{itemize}

We have +1 adjustment in weighting factor \begin{math}\frac{year-2002+1}{2023-2002+1}\end{math}. This ensures that cases from the year 2002 are still included in the weighted calculation and do not receive a weight of zero.

We did not use more than four bins for the Citation-Label to ensure enough examples also for the smallest class while still adding more granularity compared to the LD-Label.

\section{Dataset Example}

\begin{figure*}[ht]
    \centering
    \footnotesize
    \noindent\fbox{
    \parbox{\textwidth} {

    \begin{minipage}[t]{0.74\textwidth}
        \textbf{Input}\\
        \tiny{
           [Consideraions]: Erw\"{a}gungen:
        1. Angefochten ist der in einem kantonal letztinstanzlichen Scheidungsurteil festgesetzte nacheheliche Unterhalt in einem Fr. 30'000.-- \"{u}bersteigenden Umfang; auf die Beschwerde ist somit einzutreten (Art. 72 Abs. 1, Art. 74 Abs. 1 lit. b, Art. 75 Abs. 1 und Art. 90 BGG).
        2. Die Parteien pflegten eine klassische Rollenteilung, bei der die Ehefrau die Kinder grosszog und sich um den Haushalt k\"{u}mmerte. Infolge der Trennung nahm sie im November 2005 wieder eine Arbeitst\"{a}tigkeit auf und erzielt mit einem 80\%-Pensum Fr. 2'955.-- netto pro Monat. Beide kantonalen Instanzen haben ihr jedoch auf der Basis einer Vollzeitstelle ein hypothetisches Einkommen von Fr. 3'690.-- angerechnet. Das Obergericht hat zwar festgehalten, der Ehefrau sei eine Ausdehnung der Arbeitst\"{a}tigkeit kaum m\"{o}glich, gleichzeitig aber erwogen, es sei nicht ersichtlich, weshalb sie nicht einer Vollzeitbesch\"{a}ftigung nachgehen k\"{o}nne. Ungeachtet dieses Widerspruches wird das Einkommen von Fr. 3'690.-- von der Ehefrau ausdr\"{u}cklich anerkannt, weshalb den nachfolgenden rechtlichen Ausf\"{u}hrungen dieser Betrag zugrunde zu legen ist. [...]
        }
    \end{minipage}
    \hfill
    \begin{minipage}[t]{0.22\textwidth}
        \textbf{Metadata:} \\
        \tiny{
            Decision ID: \textit{65aad3f6-33c2-4de2-91c7-436e8143d6ea}\\ Year: \textit{2007}\\
            Language: \textit{German}\\
            Law Area: \textit{Civil}\\
            LD Label: \textit{Critical}\\
            Citation Label: \textit{Citation-1}\\
            Court: \textit{CH\_BGer}\\
            Chamber: \textit{CH\_BGer\_005,}\\
            Canton: \textit{CH}\\
            Region: \textit{Federation}
        }
    \end{minipage}
    \rule{\textwidth}{0.4pt}

    \begin{minipage}[t]{\textwidth}
        \textbf{Target}:\\
        \tiny{\textbf{critical-1}}\\
        \tiny{
            {Possible LD label: critical, non-critical, Possible citation label: critical-1, critical-2, critical-3, critical-4}
        }\\
        \rule{\textwidth}{0.4pt}
    \end{minipage}
   

    }
    }
\caption{Example of an input, target and corresponding metadata}
\label{tab:CP_task}
\end{figure*}

\Cref{tab:CP_task} shows an example input, target and corresponding metadata.


\section{General Dataset Metadata}

\Cref{tab:metadata} shows metadata for cantons, courts, chambers law-areas and languages.

\begin{table*}[ht]
    \centering
    \caption{Metadata for cantons, courts, chambers, law-areas, and languages}
    \footnotesize
    \resizebox{\textwidth}{!}{%
    \begin{tabular}{lll}
         \toprule
         \textbf{Metadata} &  \textbf{Number}&  \textbf{Examples}\\
         \midrule
            Cantons  & 1& Federation (CH)\\
            Courts  & 1 & Supreme Court \\
            Chambers  & 13 &  
CH-BGer-011
CH-BGer-004
CH-BGer-008
CH-BGer-002
CH-BGer-005
CH-BGer-001
CH-BGer-006
CH-BGer-009
CH-BGer-015 ...\\
            Law-Areas  & 4 & Civil, Criminal, Public, Social\\
            Languages  & 5 & German, French, Italian \\
        \bottomrule
    \end{tabular}
    }
    \label{tab:metadata}
    \vspace{-2ex}
\end{table*}

\section{Dataset Licensing}

The original case data is available from the Swiss Federal Supreme Court\footnote{\url{https://www.bger.ch/de/index.htm}} and the Entscheidsuche portal\footnote{\url{https://entscheidsuche.ch/}} was used to download HTML files for each case.

In compliance with the Swiss Federal Supreme Court’s licensing policy\footnote{\url{https://www.bger.ch/files/live/sites/bger/files/pdf/de/urteilsveroeffentlichung_d.pdf}}, we are releasing the dataset under a CC-BY-4.0 license. \textit{The link to the dataset will be made available upon acceptance.}

Personally identifying information has already been anonymized by the Swiss Federal Supreme Court in accordance with its anonymization rules\footnote{\url{https://www.bger.ch/files/live/sites/bger/files/pdf/Reglemente/Anonymisierungsregeln_2020_def__d.pdf}}.

\section{Zero-shot Prompts}

\Cref{fig:CP-LD-Facts,fig:prompt-citation} show the zero-shot prompts we used for the Case Prioritization and Leading Decision Selection tasks respectively.

\begin{figure}[ht]
    \centering
    \footnotesize
    \noindent\fbox{
    \parbox{0.9\columnwidth} {
    \textbf{\acf{CP} LD Facts/Consideration}\\\\
Given the \textcolor{blue}{\textbf{\{facts/considerations\}}} from the following Swiss Federal Supreme Court Decision:\\\textit{\textcolor{teal}{\{INPUT FROM THE VALIDATION SET\}}}\\Federal Supreme Court Decisions in Switzerland that are published additionally get the label critical, those Federal Supreme Court Decisions that are not published additionally, get the label non-critical. Therefore, there are two labels to choose from:\\
- \textcolor{orange}{\textbf{critical}}\\- \textcolor{orange}{\textbf{non-critical}}\\The relevant label in this case is:
    }
    }
\caption{Prompt used for Criticality Prediction (LD Facts/Consideration). The LLM is tasked with predicting whether a Swiss Federal Supreme Court decision is labeled as critical or non-critical based on the provided facts or considerations}
\label{fig:CP-LD-Facts}
\end{figure}

\begin{figure}[ht]
    \centering
    \footnotesize
    \noindent\fbox{
    \parbox{0.9\columnwidth} {
\textbf{\acf{CP} Citation Facts/Consideration}\\
Given the \textcolor{blue}{\textbf{\{facts/considerations\}}} from the following Swiss Federal Supreme Court Decision:\\\textit{\textcolor{teal}{\{INPUT FROM THE VALIDATION SET\}}}\\How likely is it that this Swiss Federal Supreme Court Decision gets cited. Choose between one of the following labels (a bigger number in the label means that the court decision is more likely to be cited):\\
- \textcolor{orange}{\textbf{critical-1}}\\- \textcolor{orange}{\textbf{critical-2}}\\- \textcolor{orange}{\textbf{critical-3}}\\- \textcolor{orange}{\textbf{critical-4}}\\The relevant label in this case is:
    }
    }
\caption{ Prompt used for Criticality Prediction (Citation Facts/Consideration). The LLM is tasked with predicting how likely a Swiss Federal Supreme Court decision is to be cited, using a four-tiered label (critical-1 to critical-4), based on the provided facts or considerations.}
\label{fig:prompt-citation}
\end{figure}

\section{Hyperparameters and Package Settings}

We used a fixed learning rate of 1e-5 without tuning, running each experiment with three random seeds (1-3) and excluding seeds with high evaluation losses. Gradient accumulation was applied when GPU memory was insufficient to maintain a final batch size of 64. Training employed early stopping with a patience of 5 epochs, based on validation loss. To reduce costs, AMP mixed precision was used where it didn't cause overflows (e.g., mDeBERTa-v3). The max-sequence length was set at 2048 for Facts and 4096 for Considerations. For LLM evaluations, we used the \href{https://platform.openai.com/docs/guides/gpt/chat-completions-api}{ChatCompletion API} for GPT-3.5 (as of June 7, 2023), 
and ran LLaMA-2 locally with 4-bit quantization.

For the analysis of consideration and fact lengths, we used SpaCy’s \textit{en\_core\_web\_sm} for tokenization.

\section{Resources}
The experiments were run on NVIDIA GPUs, including the 24GB RTX3090, 32GB V100, 48GB A6000, and 80GB A100, using approximately 50 GPU days in total.

\section{Additional Results}

\Cref{tab:config_aggregate_scores_with_std} shows results with standard deviations on the test set. \Cref{tab:config_aggregate_scores_of_validation_set} shows scores on the validation set.



\begin{table}[ht]
    \centering
    \caption{Configuration aggregate scores with standard deviations on the test set. The macro-F1 scores are provided.}
    \footnotesize
    \resizebox{\columnwidth}{!}{%
    \begin{tabular}{lllllllllr}
\toprule
                \bf Model &                      \bf LD-F &                      \bf LD-C &                      \bf C-F &                      \bf C-C &  \bf Agg. \\
\midrule
               MiniLM & 54.7\textsubscript{+/-1.9} & 65.8\textsubscript{+/-1.6} &  9.8\textsubscript{+/-2.8} & 20.8\textsubscript{+/-3.0} &  21.8 \\
           DistilmBERT & 56.2\textsubscript{+/-0.5} & 65.4\textsubscript{+/-1.7} & 19.6\textsubscript{+/-1.1} & 22.1\textsubscript{+/-0.4} & 30.9\\
          mDeBERTa-v3 & 55.1\textsubscript{+/-2.0} & 69.8\textsubscript{+/-2.8} & 21.0\textsubscript{+/-3.6} & 17.5\textsubscript{+/-4.4} & 29.1 \\
           XLM-R\textsubscript{Base} & 57.2\textsubscript{+/-1.5} & 65.9\textsubscript{+/-3.2} & 21.3\textsubscript{+/-1.5} & 23.7\textsubscript{+/-1.9} & 32.8 \\
          XLM-R\textsubscript{Large} & 56.4\textsubscript{+/-1.8} & 67.9\textsubscript{+/-1.9} & 24.4\textsubscript{+/-7.2} & 29.1\textsubscript{+/-2.7} &  37.1 \\
           X-MOD\textsubscript{Base} & 56.6\textsubscript{+/-1.8} & 67.8\textsubscript{+/-2.9} & 20.0\textsubscript{+/-3.0} & 20.6\textsubscript{+/-3.5} & 30.5 \\
    SwissBERT\textsubscript{(xlm-vocab)} & 56.9\textsubscript{+/-0.7} & 67.3\textsubscript{+/-4.7} & 25.7\textsubscript{+/-8.3} & 23.0\textsubscript{+/-4.0} &  34.8 \\
    \midrule
     mT5\textsubscript{Small} & 52.2\textsubscript{+/-1.9} & 62.1\textsubscript{+/-5.2} & 13.2\textsubscript{+/-2.4} & 17.9\textsubscript{+/-1.7} & 24.0\\
      mT5\textsubscript{Base} & 52.1\textsubscript{+/-1.6} & 61.5\textsubscript{+/-3.9} & 14.0\textsubscript{+/-2.8} & 19.7\textsubscript{+/-1.6} & 25.4 \\
    \midrule
           BLOOM-560m & 53.0\textsubscript{+/-1.7} & 61.7\textsubscript{+/-4.1} & 10.7\textsubscript{+/-3.7} &  8.0\textsubscript{+/-3.5} & 15.8 \\
    \midrule
      Legal-Swiss-R\textsubscript{Base} & 57.7\textsubscript{+/-1.6} & 70.5\textsubscript{+/-2.3} & 16.2\textsubscript{+/-5.8} & 20.1\textsubscript{+/-5.6} & 28.0\\
     Legal-Swiss-R\textsubscript{Large} & 55.9\textsubscript{+/-2.2} & 68.9\textsubscript{+/-2.1} & 25.8\textsubscript{+/-7.8} & 16.3\textsubscript{+/-8.7} & 30.2\\
     Legal-Swiss-LF\textsubscript{Base} & 58.1\textsubscript{+/-2.1} & 70.8\textsubscript{+/-2.9} & 21.4\textsubscript{+/-2.9} & 17.4\textsubscript{+/-8.6} &  29.5\\
\bottomrule
\end{tabular}

    }
    \label{tab:config_aggregate_scores_with_std}
\end{table}


\begin{table}[ht]
    \centering
    \caption{Configuration aggregate scores on the validation set. The macro-F1 scores are provided. The highest values are in bold. It is important to note that the scores presented here are calculated as the harmonic mean over multiple seeds.}
    \footnotesize
    \resizebox{\columnwidth}{!}{%
    \begin{tabular}{lrrrrrrrrrrrrrrrrrrrrrrrrr}
\toprule
           \bf{Model} &                 \bf{LD-F} &                 \bf{LD-C} &                 \bf{C-F} &                 \bf{C-C} &                  \bf{Agg.} \\
\midrule
               MiniLM                &       59.1 &       71.0 &       14.9 &       36.9 &       31.9 \\
DistilmBERT            &       59.6 &       70.1 &       26.3 &       35.8 &       41.2 \\
mDeBERTa-v3           &       60.1 &       73.0 &       30.4 &       36.0 &       44.0 \\
XLM-R\textsubscript{Base}            &       60.1 &       70.5 &       26.9 &       38.5 &       42.6 \\
XLM-R\textsubscript{Large}           &       60.5 &       71.7 &       27.2 &       39.7  &       43.3 \\
X-MOD\textsubscript{Base}            &       57.1 &       71.0 &       27.0 &       33.4  &       40.6 \\
SwissBERT\textsubscript{(xlm-vocab)} &       59.0 &       72.1 &       29.4 &       38.8&       44.1 \\
\midrule
mT5\textsubscript{Small}      &       54.8 &       66.1 &       26.3 &       32.5 &       39.2 \\
mT5\textsubscript{Base}       &       55.7 &       64.4 &       24.3 &       29.3 &       36.8 \\
\midrule
BLOOM-560m            &       52.2 &       64.3 &       20.1 &       21.8 &       30.7 \\
\midrule
Legal-Swiss-R\textsubscript{Base}       &       61.2 &  \bf{73.6} &       27.7 &       41.0 &  44.2 \\
Legal-Swiss-R\textsubscript{Large}      &  \bf{61.8} &       73.5 &       29.8 &       32.0 &       42.3 \\
Legal-Swiss-LF\textsubscript{Base}      &       59.4 &       72.7 &  \bf{32.2} &  \bf{42.5} &  \bf{47.0} \\
\bottomrule
\end{tabular}

    }
    \label{tab:config_aggregate_scores_of_validation_set}
\end{table}

\section{Use of AI Assistants}
We used ChatGPT and Claude to enhance grammatical correctness and style, and utilized Google Colab's Generate AI feature for some of the dataset analysis.

\end{document}